\documentclass[11pt,a4paper]{article}
\usepackage[hyperref]{eacl2021}
\usepackage{times}
\usepackage{latexsym}
\usepackage{graphicx}
\usepackage{arydshln}
\usepackage{subcaption}
\usepackage{tikz}
\usepackage{tikz-dependency}
\usepackage[T1]{fontenc}
\usepackage{amsmath}
\usepackage{centernot}
\usepackage{amssymb}
\usepackage{enumitem}

\usepackage{cleveref}

\crefformat{section}{\S#2#1#3} 
\crefformat{subsection}{\S#2#1#3}
\crefformat{subsubsection}{\S#2#1#3}

\DeclareMathOperator*{\argmax}{argmax}

\usepackage{microtype}

\aclfinalcopy 
\def\aclpaperid{667} 
\usepackage[show]{notes}
\newcommand{\ignore}[1]{}
\newcommand{\com}[1]{}

  {\begin{list}{}%
          {\setlength{\leftmargin}{#1}}%
          \item[]%
  }
  {\end{list}}

\usepackage{xspace}
\newcommand{\opt}{\texttt{B-BOC}\xspace}

\title{With Measured Words: Simple Sentence Selection for Black-Box Optimization of Sentence Compression Algorithms}

\author{Yotam Shichel \\
    BenGurion University \\
  \texttt{yotamsh@post.bgu.ac.il} \And
  Meir Kalech \\
        Ben Gurion University \\
  \texttt{kalech@bgu.ac.il} \AND
  Oren Tsur \\
        Ben Gurion University \\
  \texttt{orentsur@bgu.ac.il} }

\date{}

\begin{document}
\maketitle

\begin{abstract}
Sentence Compression is the task of generating a shorter, yet grammatical version of a given sentence, preserving the essence of the original sentence. 
This paper proposes a Black-Box Optimizer for Compression (\opt): given a black-box compression algorithm and assuming not all sentences need be compressed -- find the best candidates for compression in order to maximize both compression rate and quality.
Given a required compression ratio, we consider two scenarios: (i) single-sentence compression, and (ii) sentences-sequence compression. In the first scenario, our optimizer is trained to predict how well each sentence could be compressed while meeting the specified ratio requirement. In the latter, the desired compression ratio is applied to a sequence of sentences (e.g., a paragraph) as a whole, rather than on each individual sentence. To achieve that, we use \opt to assign an optimal compression ratio to each sentence, then cast it as a Knapsack problem, which we solve using bounded dynamic programming. 
We evaluate \opt on both scenarios on three datasets, demonstrating that our optimizer improves both accuracy and Rouge-F1-score compared to direct application of other compression algorithms. 
\end{abstract}

\section{Introduction}
\label{sec:Intro}
{\em Sentence Compression} is the task of generating a short, accurate, and fluent sentence that preserves the essence of a given original sentence by removing nonessential words and/or rephrasing it in a compact form. 
Compression can take many forms, ranging from Extractive and Abstractive Summarization \citep{jing2000sentence,madnani2007multiple,cohn2008sentence,cohn2009sentence,galanis2010extractive,rush2015neural,chopra2016abstractive} to Text Simplification and Paraphrasing \citep{bannard2005paraphrasing,xu2012paraphrasing,klerke-etal-2016-improving,narayan2017split,aharoni2018split,botha2018learning}, among others. 

On the sentential level, compression is often viewed as a {\em word deletion} task \citep{knight2000statistics,knight2002summarization,filippova2008dependency,filippova2015sentence,wang2016sentence,wang2017can,zhou-rush-2019-simple}.
However, not all sentences could, or should be compressed as part of compressing a longer text they reside in. 
Consider the familiar scenario in which a full paragraph needs to be compressed in order to have an EACL paper meet the page restriction specified in the submission guidelines. A common approach by \LaTeX~ users is to first identify paragraphs ending with a short line, (e.g., this very paragraph),
then choose one or more sentences that could be compressed with a minimal loss of information -- shaving the extra line. We propose a Black-Box Optimizer for Compression (\opt) that mitigates this problem. 
Given a compression algorithm $A$, a desired compression ratio, and a document $D$, \opt chooses the best sentences to compress using  $A$ in order to produce a shorter version of $D$,  while keeping the other sentences of $D$ untouched.  \opt achieves that without explicit knowledge of the inner-workings of the given compression algorithm, hence we call it a \emph{black-box} optimizer. Selected sentences are expected to be the best candidates for compression -- balancing compression rate with compression quality.

This paper addresses two main research questions:
(1) How to predict the compression performance (preserving meaning and grammar) of an algorithm on a given sentence? (2) Given a document and a required compression ratio, how to choose the optimal \emph{subset} of sentences to compress, along with the appropriate compression ratio per each of the sentences, so that the \emph{total} compression meets the required compression requirement?

Given a gold set of pairs of sentences and their compressions, we represent each sentence as a vector of shallow and syntactic features, and train a regression model to predict its expected compression rate. \opt ranks all sentences by the predicted compression potential while considering a required compression ratio.

The document-level task could be modeled as a Knapsack optimization problem, considering the subset of sentences to be compressed in order to satisfy the overall compression requirement (capacity), with a minimal loss of information (value). The solution space covers the trade-off between aggressively compressing only a few sentences and applying minimal compression on a larger number of sentences. While the general Knapsack is NP-complete, the 0-1 variation can be approximated efficiently by using Dynamic Programming \cite{hristakeva2005different}. 

We evaluate \opt on three benchmarks commonly used for the sentence compression task. We show that applying \opt on top of state-of-the-art sentence compression models improves the performance for any desired compression rate. In addition, optimizing the \opt-Knapsack achieves top performance on the document-level task.

\section{Related Work} \label{related_work}
Early sentence compression works employ the noisy channel model, learning the words and clauses to be pruned  \cite{knight2000statistics,knight2002summarization,filippova2008dependency,clarke2008global,cohn2009sentence}. 

The top-performing sentence compression models use a Policy Network coupled with a Syntactic Language Model (bi-LSTM) evaluator \cite{zhou-rush-2019-simple}, and a stacked LSTM with dropout layers \citep{filippova2015sentence}. An extension of Filippova et al., adding syntactic features and using Integer Linear Programming (ILP), yields improved results in a cross-domain setting \cite{wang2017can}.

Sentence selection is used for document extractive summarization -- a task conceptually close to ours, in which full sentences are extracted from a long document, see \cite{nenkova2011automatic} for an overview. State-of-the-art selection is achieved by combining sentence and document encoders (CNN and LSTM) with a sentence extraction model (LSTM) and a reinforcement layer \citep{narayan2018ranking}.  

Sentence rephrasing is an abstractive approach to rewrite a sentence into a shorter form using some words that may not appear in the original sentence.
A data-driven approach to abstractive sentence summarization is suggested in \cite{rush2015neural,chopra2016abstractive}, using about four million title-article pairs from the Gigaword corpus for training, and uses a convolutional neural network model to encode the source and produce a single representation for the entire input sentence. Tree-to-tree grammar extraction method for the rewriting task is used in \cite{cohn2008sentence,cohn2009sentence}. State-of-the-art performance on the abstractive summarization task is obtained using Hierarchical Attentional Seq2Seq Recurrent Neural Network \cite{nallapati2016abstractive, see2017get}. 

\section{Task Definitions and Methodology}
\label{sec:problem_definition}
In this section we formally define the sentence-level and the document-level tasks (\cref{subsec:definition}) and provide a detailed description of the application of \opt in both settings (\cref{subsec:methodology}).

\subsection{Problem Definitions}  
\label{subsec:definition}

\paragraph{Sentence-Level Compression} Given a set of sentences $S = \{s_{i}\}^{n}_{i=1}$; a desired compression rate $\gamma$; the number of sentences to compress $k \leq n$; a compression algorithm $A$; and an oracle $\mathbb{R}:(A,S) \to [0,1]$, returning a score reflecting the compression quality (grammaticality and minimal loss of information)  $A$ would achieve on $s \in S$ --  we would like to choose a set $S^{k,\gamma} \subseteq S$ of $k$ sentences:\\
$S^{k,\gamma} =  \{s_j | \frac{|A(s_j)|}{|s_j|} \leq \gamma \land \argmax\limits_{s_j} \mathbb{R} (A,s_j) \}_{j=1}^k$. 
We call this \emph{sentence-level} compression since each sentence should meet the $\gamma$ constraint independently. 
It is important to note that $\gamma' \leq \gamma \centernot \implies S^{k,\gamma} \subseteq S^{k,\gamma'}$, since different sentences may be better compressed to different $\gamma$ values. Consider the following two sentences used to illustrate the importance of the Oxford comma: $S=$\{``\emph{I had a yummy dinner with my parents, Batman and Catwoman'',  ``I had a yummy dinner with my parents, Batman, and Catwoman}''\}\footnote{The first sentence, without the Oxford comma, implies that Batman and Catwoman are the speaker's parents, the second sentence implies that the speaker had dinner with four people -- her parents \emph{and} Batman \emph{and} Catwoman.}, and $k=1$. The first sentence could be compressed to ``I had a yummy dinner with my parents'' with a minimal loss of information, while it does not make sense to compress the second sentence this way and it should be compressed to ``I had a yummy dinner'', thus specifying $k=1$, the sentence to be compressed with minimal loss of meaning depends on the desired $\gamma$ value.

\paragraph{Document-Level Compression} In this setting, we are given a sequence of sentences $D = \{s_i\}^{n}_{i=1}$ (a paragraph or a full document), and a desired compression rate $\gamma$ that should be applied to $D$ as a \emph{whole}. That is, we wish to find an optimal subset of $S^\gamma$ that satisfies: 
\begin{align*}
\tag{1}
\argmax\limits_{S^\gamma \subseteq D} \sum_{s_i \in S^\gamma} \mathbb{R}(A,s_i) \qquad \qquad \qquad \qquad \quad  \\
\text{s.t.} \quad  \frac{\sum_{s_i \in S^\gamma} |A(s_i)| + \sum_{s_i \in D \setminus S^\gamma} |s_i| }{|D|} \leq \gamma  
\label{eq:doc_level_def}
\end{align*}

Since $\gamma$ refers to $D$ rather than to individual sentences, the overall quality can be maximized by choosing a varying number of sentences expected to achieve different optimal compressions. Unlike the sentence-level setting, here, an optimal $S^\gamma$ may contain a combination of sentences, for some of which $\frac{|A(s)|}{|s|} \leq \gamma$, and for others  $\frac{|A(s)|}{|s|} > \gamma$.

\subsection{Computational Approach} 
\label{subsec:methodology}

\paragraph{Scoring Function}
Given a corpus $C = \{\langle s_i, \hat{s_i}\rangle\}_i^m$ of sentence pairs, each pair contains an original sentence $s$ and its gold compression $\hat{s}$, we define the golden ratio $\hat{\gamma_i} = \frac{|\hat{s_i}|}{|s_i|}$, and posit $\mathbb{R} (A,s) \approx 1- | \hat{\gamma_i} - \frac{|A(s_i)|}{|s_i|}|$.

We justify the use of $\hat{\gamma_i}$ as a proxy to the optimal compression quality, as compression ratios are found to correlate with compression quality measured against gold compressions \cite{napoles2011evaluating}. The use of $\hat{\gamma_i}$ as a proxy is validated through manual evaluation, see Sections \ref{subsec:eval_setup} and \ref{subsec:results}. 

Syntactic features were successfully used for sentence compression \cite{clarke2008global,wang2017can,liu2017dependency,futrell-levy-2017-noisy}. Assuming that sentence complexity correlates with the ease of compression, we follow \cite{brunato2018sentence} and represent each sentence as a vector of shallow features (sentence length, average word length, punctuation counts, etc.) and syntactic features (depth of constituent parse tree as well as the number of internal nodes, word's depth in a dependency parse tree, mean dependency distance, etc.).

We now train a regression model and learn the scoring function $R(A,s)$  by minimizing the loss:
\begin{equation*}
  L(C, A) = \Sigma_{s_i \in C}\left[R(A,s_i) - \mathbb{R} (A,s_i)\right]^2
\end{equation*}

We note that we do not train a compression algorithm, but an oracle -- a scoring function that predicts the \emph{quality} of the compression algorithm $A$ will achieve on a given sentence. This oracle will be used to rank candidate sentences in order to optimize the choice of sentences in the two tasks defined in Section \ref{subsec:definition}.

We train a Gradient Boosted Tree regression model using XGBoost. The model's hyperparameters (e.g., subsample ratio, learning rate, max depth) were tuned on a separated development set.

\paragraph{Sentence level compression:}
Given a set of sentences $S$, \opt operates on two steps: (i) It applies $R$ on every $s \in S$, producing an ordered set $\hat{S}$ for which  $\forall_{i<j} R(A,s_i) \geq R(A, s_j)$. (ii) It constructs $S^{k,\gamma}$ by iterating over $\hat{S}$, choosing the first $k$ sentences that satisfy the $\gamma$ requirement.

\paragraph{Document level compression:} Using the task definition in Section \ref{subsec:definition}, it is straight forward to cast the task as a combinatorial 0-1 Knapsack problem in the following way:
Given a set of items (sentences) $S = \{s_1,... ,s_n\}$, each weighs $w_i=|A(s_i)|$  if compressed, or $|s_i|$ if kept in the original form, and each holds a value $v_i = R(A, s_i)$ (predicted compression quality), if compressed and $v_i=1$ if kept in the original form;  and given a weight limit $W = \gamma \cdot \sum_{1}^n |s_i| $
 -- we wish to find $S^\gamma = \{s_i| x_i=1\}$ that maximizes:

\begin{align*}
\qquad \sum_i (v_i x_i- \left[1-x_i\right]^{-1}) \qquad \qquad  \text{s.t.} \quad \qquad \\
\sum_i (w_i x_i-|s_i| \left[1-x_i\right]^{-1}) \leq W \quad, x_i \in \{0,1\}
\label{eq:doc_level_knap}
\end{align*}

were $x_i=1$ denotes we choose to compress $s_i$ and $x_i=0$ denotes that $s_i$ remains in its original long form (hence the 0-1 Knapsack setting). Note that the value we maximize and the weight constraints include a term for the unchanged sentences, in case they are not chosen for compression. This term is introduced since the $\gamma$ constraint in the task definition applies to the document as a whole. 

\opt-knapsack returns $S^\gamma$ by solving the 0-1 knapsack problem using the dynamic programming approach proposed by \cite{hristakeva2005different} to reduce the computation complexity to a pseudo-polynomial time. Knapsack's solution ensures an optimal set of sentences, satisfying the required compression limitations, while achieving the maximum quality score.

\section{Experimental Setting}
\label{sec:evalualtion}

\subsection{Datasets}
\label{subsec:datasets}

\paragraph{Training Data:} \label{sec:train_data}
We train \opt on a dataset of 200,000 sentence-compression pairs\footnote{\url{www.github.com/google-research-datasets/sentence-compression}} used by Filippova and Altun \shortcite{filippova-altun-2013-overcoming}. Each pair is composed of a long sentence (usually the teaser, caption, extract or the first sentence that bears the most salient information) from a news story and the story's  headline, which is a compressed version of the long sentence.

Out of these 200,000 sentences, we set aside 9,000 to be used as a development set, and 1,000 as one of our three test sets. 

\paragraph{Evaluation datasets:}  \label{sec:datasets}
Three datasets are used for evaluation: 
\begin{enumerate}[noitemsep]
    \item Google (GGL) -- the first 1000 sentences of the training corpus (described above) were used for testing.
    \item British National Corpus (BNC) -- a manually crafted dataset of  $\sim 1500$ sentence-compression pairs. Given a long sentence, annotators were asked to produce a short version by deleting extraneous words from the source without changing the order of words \footnote{\url{jamesclarke.net/research/resources}}.
    \item Gigaword (GIGA)- headline-generation corpus of articles\footnote{\url{github.com/harvardnlp/sent-summary}} consists $	\sim4$ million sentence-compression pairs. We note that this dataset contains abstractive pairs, nevertheless, it can be used to measure accuracy.
\end{enumerate}

\subsection{Evaluation Procedures} \label{subsec:eval_setup}

\paragraph{Evaluation metrics:}  \label{sec:eval_metrics} 
We used four evaluation metrics that complement each other, providing a comprehensive evaluation of the different factors that contribute to quality summarization as suggested by \citep{filippova2015sentence}: (1) Accuracy -- how many compressed sentences are fully reproduced, (i.e., the generated compression is identical to the golden one). (2) F-score -- given the golden and predicted compressions, recall and precision are based on the ROUGE metric. (3) Readability score -- the grammaticality of the compression. (4) Informativeness -- the level in which the compression covers the most salient information. 

The two latter metrics are based on a manual evaluation by three annotators, scoring Readability and Informativeness on a 5-Point Likert scale. The annotators were guided to give a top Readability score (score 5) if the predicted compressed sentence is clear and grammatically correct, regardless of the original context, and a top Informativeness score (score 5) if the essence of the original is preserved completely. The Informativeness measurement bears some degree of subjectivity as annotators may not agree on what should be considered ``the essence'' of a sentence, see examples in Table \ref{tbl:sentences_example}. We used Cohen's Kappa \cite{cohen1960coefficient} to measure inter-annotator agreements. Low agreements are expected due to the subjectivity and the five-point scale, i.e., when two raters agree on the grammaticality of a sentence, but do not give the same exact Informativeness score. To account for slight variations in assessment, we measure agreement using the off-by-one procedure proposed by \cite{tsur2009revrank} and supported by \cite{toutanova-etal-2016-dataset}. Linear and Quadratic weighting were added as additional statistical methods. Nevertheless, we kept the 5-point scale to be aligned with Filippova's evaluations. The Kappa values for the strict and the off-by-one agreement for a sample of 200 sentences of the GGL dataset are reported in Table \ref{tbl:Kappa_agreement}. These scores are comparable  with the scores reported by \citep{filippova2015sentence}. Agreement of 0.86 and 0.78 for Readability and Informativeness reflect an almost perfect agreement on Readability and substantial agreement on Informativeness, according to the interpretation protocol suggested by  \citet{mchugh2012interrater}.

\begin{table*}
\small
\begin{tabular}{c|p{9cm}|p{4.5cm}}
 Source  & Text & Issue \\
 \hline
Long & A gang of youths between eight and sixteen robbed a man in an Oldbrook underpass for just 10\pounds & The salience the clause  ``for just 10\pounds'' \\
Manual 1 &  \textit{A gang of youths robbed a man in an Oldbrook underpass for just 10\pounds}&\\
Manual 2&  \textit{A gang robbed a man in an Oldbrook underpass}&\\
\hline
Long & A woman was injured by a falling tree in the Gresham neighborhood, according to the Chicago Fire Department& The salience of the location  ``Gresham neighborhood'' \\
Manual 1 & \textit{A woman was injured by a falling tree} & \\
Manual 2 & \textit{A woman was injured by a falling tree in the Gresham neighborhood} & \\
\hline
\end{tabular}
\caption{Two examples of compression disagreements. }
\label{tbl:sentences_example}
\end{table*}

\begin{table}
\begin{center}
\resizebox{6cm}{!}{
\begin{tabular}{p{1.8cm}||c|c}
Agreement coverage&Readability&Informativeness\\
 \hline
Strict & 0.61 & 0.32\\
Off-by-one & 0.86 & 0.78\\
Linear & 0.78 & 0.54\\
Quadratic & 0.87 & 0.72\\
\end{tabular}}
\end{center}
\caption{Cohen's Kappa inter-annotator agreement between three annotators for the strict, off-by-one, and other statistical approaches to calculate agreement.}
\label{tbl:Kappa_agreement}
\end{table}

\subsubsection{Black-Box Compression Models} \label{sec:eval_models}
As described in Section \ref{subsec:methodology}, \opt accommodates any compression model used to compress the sentences. To show this independence, \opt is evaluated with three competitive compression models: (1) \texttt{Filippova}: An LSTM model trained on two million sentence-compression pairs \cite{filippova2015sentence}, (2) \texttt{Zhou}: An unsupervised model for sentence summarization \cite{zhou-rush-2019-simple}, and (3) \texttt{Klerke}: A three-layer bi-LSTM model \cite{klerke-etal-2016-improving}.

\subsection{Experimental Procedure}
Given the two settings presented in Section \ref{subsec:definition}, we aim to evaluate the performance of \opt in optimizing compression quality, on top of a number of black-box compression algorithms. We evaluate the way different values of $k$ affect the performance, and explore the contribution of various feature types to the trained optimizer. 

\label{sec:eval_metod}

\subsubsection{Sentence level compression:} \label{sec:sentence_level}
We evaluate the effectiveness of \opt for varied compression rates. The tested sentences were divided into buckets of different compression rates 0.1-0.9. For each  bucket we set $k$ to be 50\% of the sentence in a bucket and compare \opt selections to: (1) A random selection of $k$ sentences from the bucket (RANDOM). (2) The average of all sentences in the compression rate bucket (ALL). We report results of this comparison for each of the black-box algorithms listed in Section \ref{sec:eval_models}). Note that the F-score is based on the actual results of each of the black-box models, and that both \opt and RANDOM choose from the same pool of candidates for each compression rate bucket.

\subsubsection{Document level compression} \label{sec:document_level}
Having a document or a paragraph comprised of several sentences that are needed to be compressed, the target is to find the sentences that would gain the highest performance score subject to the overall compression ratio constraint.

To simulate a document, we synthesized one hundred documents by randomly sampling sentences from the test set. Every document contains 100 different sentences of varying lengths. We then use \opt-Knapsack as described in \ref{subsec:methodology}.
 \opt-Knapsack is compared with: (1) an oracle Knapsack solution where the golden scores are provided, rather than estimated by \opt (ORACLE). (2) We iteratively sample sentences to compress until the compression ratio is reached (RANDOM). (3) A sorted selection- choosing sentences by their lengths in an ascending sort (SHORTER FIRST). The latter baseline was added followed by our experiments, showing that compression quality tends to be higher for shorter sentences.

\section{Results and Discussion} 
\label{sec:eval_results}

\subsection{Results}
\label{subsec:results}
Detailed results for both sentence level and document level compression are presented below.

\begin{figure*}
\centering
\begin{subfigure}{.5\textwidth}
  \centering
  \includegraphics[width=0.9\linewidth]{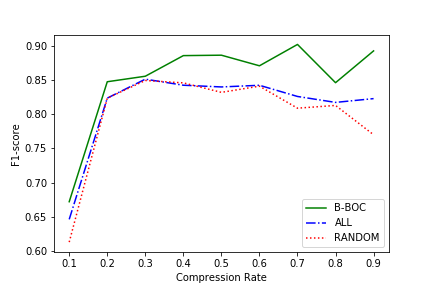}
  \caption{Filippova's compression model}
  \label{fig:FilippovaF1Fig}
\end{subfigure}%
\begin{subfigure}{.5\textwidth}
  \centering
  \includegraphics[width=0.9\linewidth]{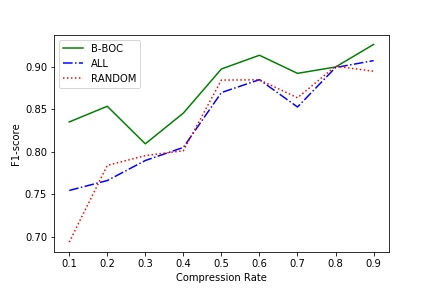}
  \caption{Zhou's compression model}
  \label{fig:ZhouFig}
\end{subfigure}
\begin{subfigure}{.5\textwidth}
  \centering
  \includegraphics[width=0.9\linewidth]{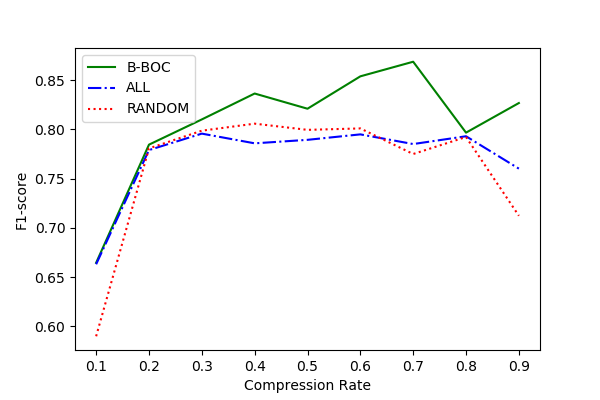}
  \caption{Klerke's compression model}
  \label{fig:ClerkeFig}
\end{subfigure}
  \caption{Average F1-score ($y$-axis) applied on the GGL dataset for different compression rate buckets ($x$-axis) while training \opt with Filippova (a), Zhou's (b) and Klerke (c) compression models.}
\label{fig:FilipZhouGGLFig}
\end{figure*}

\paragraph{Sentence level compression:} 
Figure \ref{fig:FilipZhouGGLFig} presents the F1 performance of sentence selection methods over varied $\gamma$-buckets on the GGL dataset, while training with the compression models of Filippova and Zhou respectively. \opt is compared with all sentences and a random selection of sentences, as described in Section \ref{sec:sentence_level}. It can be seen that \opt achieves the highest F1-score for every $\gamma$. 
The evaluated metrics' averages for all compression buckets are presented in Table \ref{tbl:results}, evaluating the GGL dataset using three different compression models. Best results are in bold. \opt achieves the best performance overall measures -- automatic and manual (F1-score, Accuracy, Readability and Informativeness).

\begin{table}
\begin{center}
\resizebox{8cm}{!}{
\begin{tabular}{l|l||c|c|c|c|c}
& & F1-score & F2-score & Accuracy & Readability & Info.\\
 \hline
&ALL&0.837&0.77&0.31&4.562&3.78\\
Filippova& RANDOM&0.835&0.76&0.306&4.559&3.79\\
 &\opt&{\bf 0.86 *}&{\bf 0.795}&{\bf 0.332}&{\bf 4.65}&{\bf 4.08}\\
 \hline
&ALL&0.82&0.7&0.24&{\bf 3.92}&3.41\\
Zhou&RANDOM&0.815&0.69&0.226&3.91&3.41\\
&\opt&{\bf 0.87 *}&{\bf 0.77}&{\bf 0.30}&{\bf 3.92}&{\bf 3.60}\\
 \hline
&ALL&0.787&0.685&0.187&4.12&3.73\\
Klerke&RANDOM&0.783&0.677&0.156&4.00&3.65\\
&\opt&{\bf 0.815 *}&{\bf 0.72}&{\bf 0.214}&{\bf 4.23}&{\bf 3.97}\\
\end{tabular}
}
\end{center}
\caption{GGL dataset: Evaluation metrics' average results over all compression rate buckets.  Statistical significance using a paired T-test is indicated by *.}
\label{tbl:results}
\end{table}


\begin{table}
\begin{center}
\resizebox{7cm}{!}{
\begin{tabular}{c||c|c|c|c}
Likert score& Info. F1&Read. F1&Info. Var&Read. Var\\
 \hline
1&0.59&0.67&0.13&0.10\\
2&0.66&0.66&0.07&0.04\\
3&0.70&0.69&0.05&0.05\\
4&0.77&0.74&0.04&0.05\\
5&0.77&0.75&0.03&0.04\\
\end{tabular}
}
\end{center}
\caption{Readability and Informativeness average F1-scores and variance.}
\label{tbl:likert_avg}
\end{table}

The results confirm that by utilizing \opt, the top sentences which yield the best overall compression results will be chosen, no matter which black-box compression model is applied, for every given compression ratio. Table \ref{tbl:likert_avg} describes the average F1-scores and variances for the manual evaluations of the GGL dataset using Filippova's compression model. It can be seen that both Readability and Informativeness are correlated with F1-scores. A compression that gets a higher Readability or Informativeness on the 5-point Likert score, will most probably get a higher F1-score as well, with a lower variance. This manual evaluation of Readability and Informativeness supports our choice of $\mathbb{R}$ (see Section \ref{subsec:methodology}).

\begin{figure}
\resizebox{7.5cm}{!}{
  \includegraphics[width=1\linewidth]{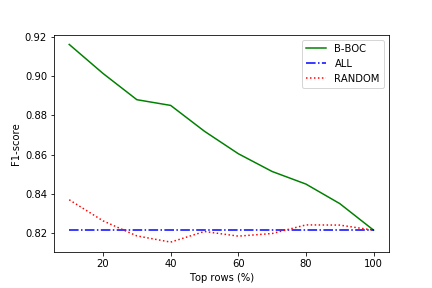}
  }
  \caption{F1-score ($y$-axis) per top X\% of sentences ($x$-axis) that are ranked by \opt.}
  \label{fig:determinizm}
  
\end{figure}

As described in Section \ref{sec:sentence_level}, the top 50\% of the test dataset are chosen for our evaluations. We repeated the same experiment varying the number (percentage) of sentences ranked by \opt. Figure \ref{fig:determinizm} presents the impact of the number of sentences we consider, and demonstrates the deterministic trend of \opt ranking method. It can be seen that when considering only the higher ranked sentences, their compression will produce a higher F1-score. It suggests that the lower number of sentences we consider -- the higher the benefit of \opt is, compared with a random selection of the same number of selected sentences. 

\begin{figure*}
\centering
\begin{subfigure}{.5\textwidth}
  \centering
  \includegraphics[width=0.95\linewidth]{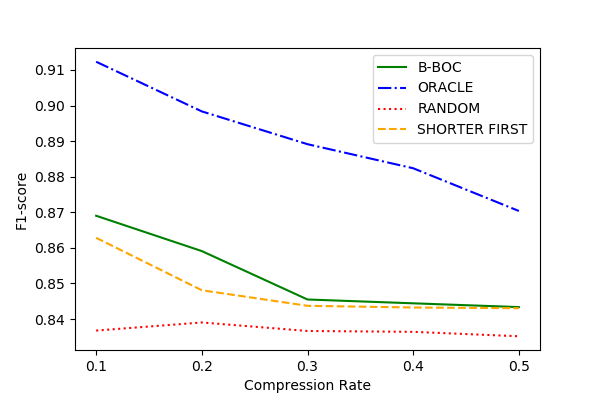}
  \caption{GGL dataset}
  \label{fig:knapsack-google}
\end{subfigure}%
\begin{subfigure}{.5\textwidth}
  \centering
  \includegraphics[width=0.95\linewidth]{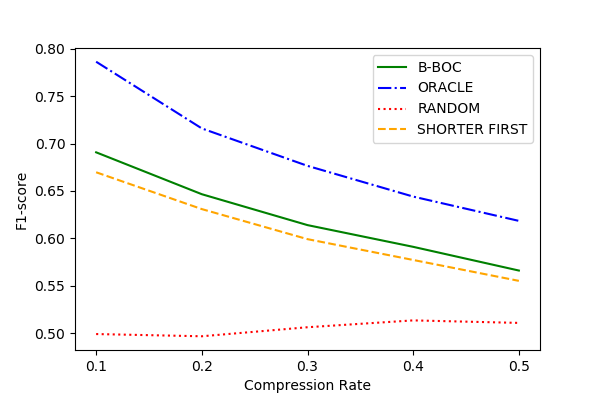}
  \caption{BNC dataset}
  \label{fig:knapsack-clarke}
\end{subfigure}
  \caption{0-1 Knapsack's F-score results for GGL dataset (a) and BNC dataset (b). $x$-axis is the total desired compression rate of the document (i.e., 0.1 means compressing the whole document by 10 percent). $y$-axis is the average F1-score of the subset of sentences being compressed.  }
\label{fig:knapsack-res}
\end{figure*}

\begin{figure}[!tbp]
\begin{center}
\resizebox{6cm}{!}{
  \centering
  \includegraphics[width=1\linewidth]{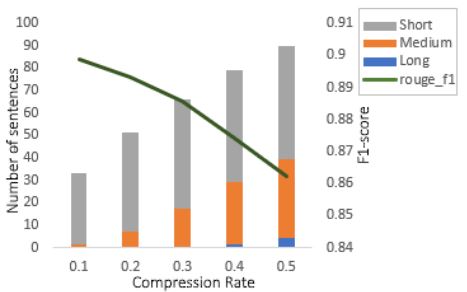}
  }
  \caption{GGL dataset: Oracle Knapsack subset's histogram. Two $y$-axis are average F1-score of the subset (right axis, describes the line) and the number of selected sentences to be compressed (left axis, describes the bars' height). $x$-axis is the total desired compression rate of the document.}
  \label{fig:GoogleKnapsackHistogram}
\end{center}
\end{figure}

\paragraph{Document level compression:}
Given a document or a paragraph and a specified compression rate requirement, \opt-Knapsack aims to find a subset of sentences, that together will satisfy the compression rate constraints if being compressed, and while guaranteeing a top F1-score. Our results below depict an experiment for compressing a document with a certain compression ratio constraint. A document is constructed using 100 sentences with variate lengths, randomly selected from a given dataset. \opt-Knapsack sentence selection is being compared with a random selection and a sorted selection of the sentences, as described in Section \ref{sec:document_level}. Each experiment was repeated 100 times, sampling different sentences for each of the datasets. The average scores reported below where achieved  with the same compression model used by Filippova (see Section \ref{sec:eval_models}) for all sentences.

Figure \ref{fig:knapsack-res} presents the experiments for GGL and BNC datasets respectively. An overall compression requirement is added, ranging from 0.1 to 0.5 (e.g., 0.1 means that the document should be compressed in 10 percent). \opt-Knapsack has a higher F1-score for almost every compression ratio, especially at the lower ratios. 

Knapsack's oracle solution can be created when considering the actual F1 and compression rates for all sentences. A histogram of the sentences that the oracle Knapsack chose to compress, grouped by their lengths is presented in Figure  \ref{fig:GoogleKnapsackHistogram}.
The Figure provides a number of insights: (1) The F1-score decreases as the number of compressed sentences grow, due to the increased uncertainty when compressing more sentences. A similar pattern is observed in Figure \ref{fig:knapsack-res}. (2) The Knapsack prefers to choose shorter sentences, as these perform better than longer sentences. We attribute this to the fact that shorter sentences may be easier to optimize, as compression alternatives are limited, compared to longer sentences.

The average results for the three datasets are presented in Table \ref{tbl:results-knapsack}. Best results are in bold. Informativeness and Readability average scores are aligned with the F1-scores (note that the GIGA dataset was not manually annotated for readability and informativeness, since we are focused on extractive summarization rather than abstractive, and the readability and and informativeness of the two types cannot be compared directly).We observe that \opt chooses the best sentences and provides a better compression performance for any compression ratio.

\begin{table}
\begin{center}
\resizebox{7.5cm}{!}{
\begin{tabular}{l|l||c|c|c}
& & F1-score & Readability & Informativeness\\
\hline
&ORACLE&0.89&4.64&4.05\\
\cdashline{2-5}
GGL&RANDOM&0.837&4.59&3.76\\
&SHORTER FIRST&0.850&4.6&3.97\\
&\opt&{\bf 0.854}&{\bf 4.62}&{\bf 4.01}\\
 \hline
&ORACLE&0.68&3.77&3.26\\
\cdashline{2-5}
BNC&RANDOM&0.55&3.53&3.13\\
&SHORTER FIRST&0.62&3.69&3.24\\
&\opt&{\bf 0.63}&{\bf 3.79}&{\bf 3.28}\\
 \hline
&ORACLE&0.47&-&-\\
\cdashline{2-5}
GIGA&RANDOM&0.256&-&-\\
&SHORTER FIRST&0.297&-&-\\
&\opt&{\bf 0.303}&-&-\\
\end{tabular}
}
\end{center}
\caption{Average results- Document level compression.}

\label{tbl:results-knapsack}
\end{table}

\begin{figure}
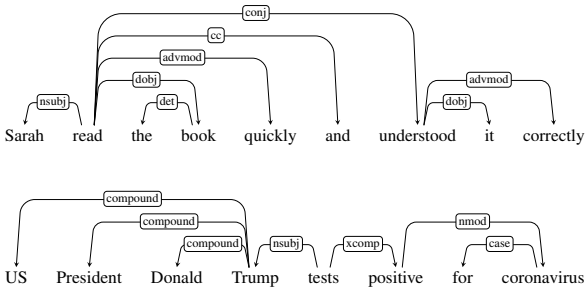

\hspace*{-.3cm}
\begin{subfigure}{.5\textwidth}
\resizebox{8cm}{!}{
\begin{dependency}
    \begin{deptext}[column sep=1.2em]
    
    Sarah \& read \& the \&  book \& quickly \&  and \&  understood \&  it \& correctly \\
    \end{deptext}
    \depedge{2}{1}{nsubj}
    \depedge{2}{4}{dobj}
    \depedge{4}{3}{det}
    \depedge{2}{5}{advmod}
    \depedge{2}{6}{cc}
    \depedge{2}{7}{conj}
    \depedge{7}{8}{dobj}
    \depedge{7}{9}{advmod}
\end{dependency}
}
  \label{fig:ADD_Score1}
\end{subfigure}

\hspace*{-.3cm}
\begin{subfigure}{.5\textwidth}
\resizebox{8cm}{!}{
\begin{dependency}
    \begin{deptext}[column sep=1.2em]
    US \& President \& Donald \& Trump \& tests \& positive \& for \& coronavirus \\
    \end{deptext}
    \depedge{5}{4}{nsubj}
    \depedge{4}{1}{compound}
    \depedge{4}{2}{compound}
    \depedge{4}{3}{compound}
    \depedge{5}{6}{xcomp}
    \depedge{6}{8}{nmod}
    \depedge{8}{7}{case}
\end{dependency}
}
  \label{fig:ADD_Score2}
\end{subfigure}
  \caption{Dependency trees of sentences of the same length (chars), but different depth and MDD score.}
  \label{fig:ADD_Score}
\end{figure}

\paragraph{Feature Importance:}
Sentence complexity is correlated with the parse tree structure \cite{oya2011syntactic}. Analyzing the contribution of each feature type, we find the tree depth features and especially Mean Dependency Distance (MDD) to do the heavy lifting.  The MDD is the sum of the depth of words in the dependency tree, divided by the total number of dependencies. For example, the MDD scores for two sentences of the same character length  ``Sarah read the book quickly and understood it correctly'' (Figure \ref{fig:ADD_Score} top) and ``US President Donald Trump tests positive for coronavirus'' (Figure \ref{fig:ADD_Score} bottom) is $19 / 8 = 2.735$ and $11 / 7 = 1.57$, respectively. 
This observation validates our intuition about the relation between sentence complexity and compression.

  Table \ref{tbl:importance} presents the importance of the syntactic features to the \opt model in terms of weight, which means the relative number of times a feature occurs in the boosted trees of the trained model. Shallow properties such as the number of verbs and number of nodes are located at the bottom.

\begin{table}
\begin{center}
\resizebox{6.5cm}{!}{
\begin{tabular}{ p{5cm}|p{2cm}}
 Feature name&Weight\\
 \hline
Word Length Average&0.140630\\
MDD-score&0.121428\\
Tree Depth Average&0.116776\\
Character count&0.093331\\
Count Relations&0.079562\\
Count PoS&0.072715\\
Parse-Tree Height&0.069478\\
Count Nodes&0.066835\\
Parse-Tree sub trees&0.061700\\
Count words&0.051541\\
Dependency tree depth&0.045512\\
Verb count&0.042461\\
Parse-Tree count POS types&0.038032
\end{tabular}}
\end{center}
\caption{Feature Importance of \opt. The percentages representing the relative number of times a particular feature occurs in the trees of the model.}
\label{tbl:importance}
\end{table}

\subsection{Discussion}
\label{subsec:discussion}
\paragraph{Limitation of the F1-score} 
Our main target is maximizing the F1-score, which happens to be a common approach for the sentence compression task, e.g., \cite{filippova2015sentence,zhao2018language}. Automatic evaluation metrics like the F1-score serve complementary purposes for linguistic quality evaluation rather than replacement because it is unclear whether the
improvement in F1-score necessarily indicates the improvement of linguistic
quality. Nevertheless, it was shown  that the F1-score correlates 
with human judgment \citep{napoles2011evaluating}. We manually performed additional evaluation for Readability and Informativeness to complement the evaluation based on the F1-score. For example, when applied on the document-level on the BNC dataset, \opt does not achieve the best F-score but does achieve best Readability and Informativeness scores (see  Table \ref{tbl:results-knapsack}). 

\paragraph{Fairness} Compression algorithms should be compared for similar levels of compression \citep{napoles2011evaluating}. Partitioning $S$ to  different compression rate buckets, as explained in \ref{sec:sentence_level} and can be seen in Figure \ref{fig:FilipZhouGGLFig}, is our way to ensure a fair comparison between the different compression models. 

\paragraph{Manual evaluations.} Exploring the cases in which annotators did not agree on either Readability or Informativeness, we noticed a higher likelihood for disagreement in the lower scale of both measurements, especially in these cases the original (long) sentence was convoluted or grammatically flawed.

  


\section{Conclusions}
In this paper we presented \opt - Black-Box Optimizer for Compression, a new complexity optimization method designated to the sentence compression problem. We defined the correlation between the complexity of a sentence and the chance that a black-box compression model could successfully compress it. Our optimization model is independent of the compression model used to compress the sentences and can be combined with any sentence compression model. Our evaluation on three benchmarks revealed promising results when applied to three different types of sentence compression models. We achieve top performance for a document compression problem using the \opt-Knapsack optimization implemented with a bounded Dynamic Programming technique. Our method could assist in compressing any kind of text while applying their desired compression model. Utilizing our method provides a proper guideline for which of the sentences are the most beneficial to focus on, in order to compress a given text, while yielding the best overall compression results.

For future work, we plan to construct Model-Dependant Optimization that accounts for the features of each  the compression model. This will facilitate a choice of the compression model that is the most suitable for a given sentence. 

\section*{Acknowledgments}
We would like to thank Katja Filippova and Alexander Rush for providing us with the outputs of their compression models.

\bibliography{eacl2021}
\bibliographystyle{acl_natbib}

\end{document}